\documentclass[10pt,twocolumn,letterpaper]{article}

\usepackage{btas}
\usepackage{times}
\usepackage{epsfig}
\usepackage{graphicx,dblfloatfix}
\usepackage{amsmath}
\usepackage{amssymb}
\usepackage{subfloat}
\usepackage{subfig}
\usepackage{makecell}
\usepackage{multirow}
\usepackage{fancyhdr}
\usepackage[multiple]{footmisc}

 \btasfinalcopy 

\ifbtasfinal\fi
\begin{document}

\title{Heterogeneity Aware Deep Embedding for Mobile Periocular Recognition}

\author{Rishabh Garg$^{1}$\thanks{Equal contributions by Garg and Baweja}, Yashasvi Baweja$^{1}$\footnotemark[1], Soumyadeep Ghosh$^{1}$, Richa Singh$^{1}$,  Mayank Vatsa$^{1}$, Nalini Ratha$^{2}$\\
$^{1}$IIIT-Delhi, India, $^{2}$IBM TJ Watson, USA\\
{\tt\small $^{1}$\{rishabh15076, yashasvi15116, soumyadeepg, rsingh, mayank\}@iiitd.ac.in, $^{2}$ratha@us.ibm.com}
}

\maketitle
\thispagestyle{empty}

\begin{abstract}
Mobile biometric approaches provide the convenience of secure authentication with an omnipresent technology. However, this brings an additional challenge of recognizing biometric patterns in an unconstrained environment including variations in mobile camera sensors, illumination conditions, and capture distance. To address the heterogeneous challenge, this research presents a novel heterogeneity aware loss function within a deep learning framework. The effectiveness of the proposed loss function is evaluated for periocular biometrics using the CSIP, IMP and VISOB mobile periocular databases. The results show that the proposed algorithm yields state-of-the-art results in a heterogeneous environment and improves generalizability for cross-database experiments.

\end{abstract}

\vspace{-10pt}
\section{Introduction}
Mobile devices are ubiquitous and they are used for various applications such as mobile banking, e-business and social media. These devices store confidential and critical data which if lost/stolen can cause harm to the user. Therefore, secure, convenient and fast authentication methods are required to unlock the devices. Most of the modern mobile devices rely on biometric based authentication~\cite{mavcek2016multimodal} such as face and fingerprint recognition to validate the identity of the user. However, biometric authentication on mobile devices pose several challenges. A primary challenge in acquiring the biometric data from mobile phones is that it is highly unconstrained. For touch-less sensing (e.g. capturing faces), the quality of the image can be adversely affected by factors such as variation in illumination conditions, distance from the subject, indoor/outdoor scenarios, quality of the front and back camera, and motion blur due to movement of the device/subject. Different mobile sensors for capturing biometric data pose a cross sensor matching problem, as different camera sensors have different imaging properties. This introduces heterogeneity in the captured data (e.g., indoor vs outdoor, front camera vs back camera resolution), and it makes biometric recognition on mobile devices an interesting and challenging problem. 

\begin{figure}
  \centering
  \includegraphics[width=0.96\columnwidth]{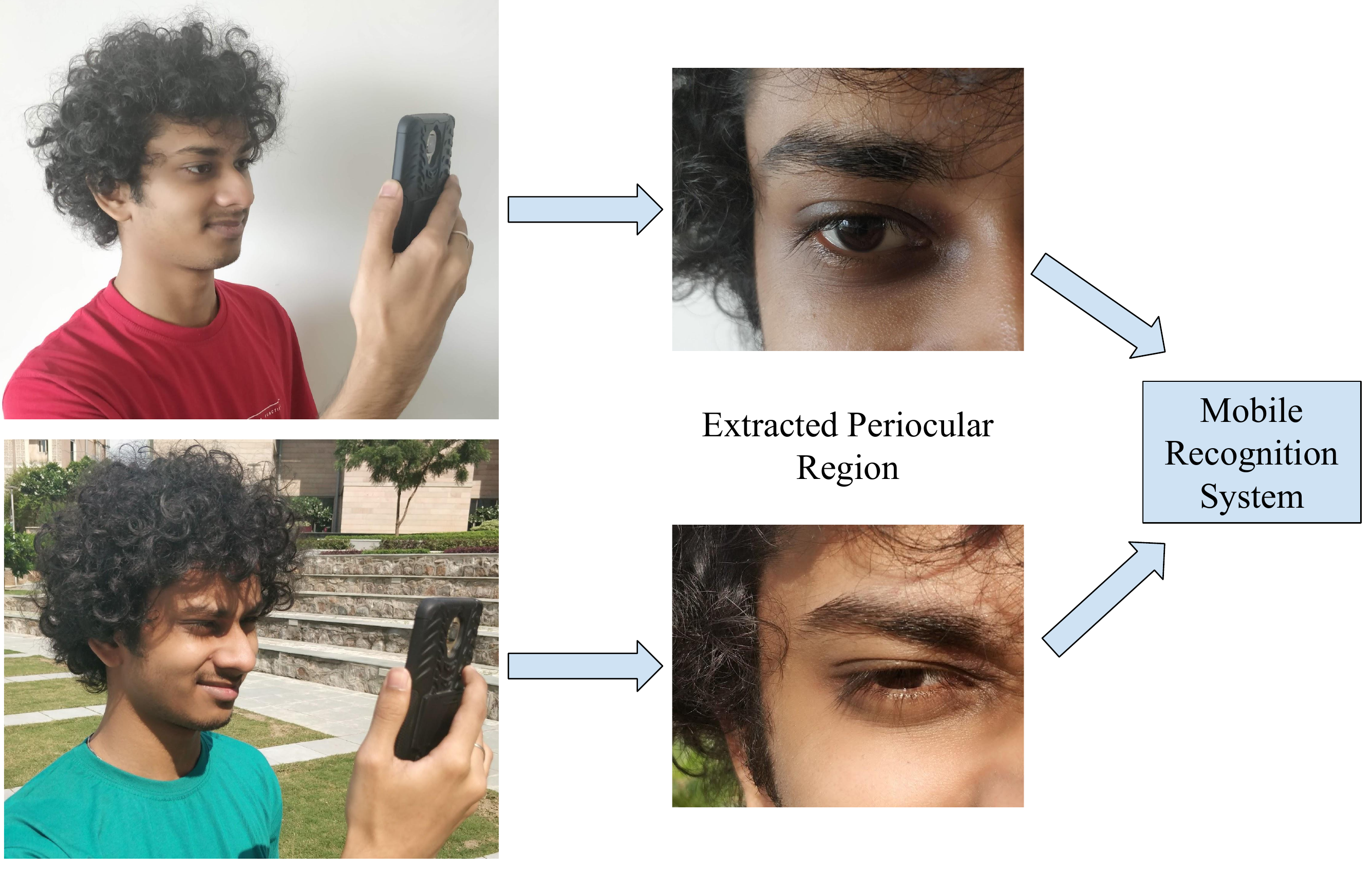}
  \caption{Data captured from mobile devices in indoor and outdoor conditions result in large variations.}
  \label{fig:mobile_eg}
\end{figure}

Periocular region as a biometric modality \cite{bharadwaj2010periocular,park2009periocular} has been gaining attention. It refers to using the regions around the eye for identity recognition. The periocular region is generally available even in unconstrained scenarios with a non cooperative subject and it can be especially useful in situations where the other information such as face is partially occluded. Figure \ref{fig:mobile_eg} illustrates the use of mobile periocular recognition in unconstrained environments. It requires no additional capturing overhead which is useful while capturing using a mobile device. 
Feasibility of periocular region as a biometric trait was explored by Park \etal \cite{park2009periocular}. Thereafter, there has been significant research advancements in this area. Detailed surveys of periocular recognition are provided by Alonso-Fernandez \etal \cite{alonso2016survey} and Nigam \etal \cite{nigam2015ocular}. A large number of techniques have performed periocular recognition on data obtained with high quality sensors in constrained conditions but there has been increasing focus on the less constrained scenarios as well.
Many popular methods relied on hand crafted features like HOG, SIFT and LBP for the periocular and iris information \cite{bharadwaj2010periocular,nigam2015ocular}. 
Tan \etal \cite{tan2013towards} use filters applied on input data for providing discriminative features for segmentation and recognition. Nie \etal \cite{nie2014periocular} use convolutional restricted Boltzmann machine along with handcrafted feature extraction for improved performance.

Recently, deep convolutional neural networks have gained immense popularity for ocular recognition. Zhao and Kumar \cite{zhao2017accurate} use explicit semantic information to extract better features and improve performance of the CNN. Proen\c ca \etal \cite{proencca2018deep}
generate artificial samples belonging to multiple classes by interchanging ocular parts from different subjects for data augmentation thereby improving the training process. Several works have also explored the problem of periocular recognition by capturing data using mobile devices. De Frietas \etal \cite{de2015periocular} model the inter session variability in the data from the enrollment time to the test time. 
Raghavendra \etal \cite{raghavendra2016learning} utilize coupled autoencoders and Maximum Response (MR) based texture features for mobile periocular recognition. Another approach by Raja \etal \cite{raja2016collaborative} used pooling of sparse filtered features. Zhang \etal \cite{zhang2018deep} use the fusion of iris and periocular region information with weighted concatenation to obtain a joint representation.


In this paper, a novel heterogeneity aware deep embedding framework for periocular recognition is proposed specifically for scenarios where the images are captured in unconstrained settings. The proposed method works by obtaining the heterogeneity invariant feature representations of the periocular images via a deep convolutional neural network. The deep CNN model is trained via the proposed heterogeneous aware loss metric based on the identity of the subjects and tries to enforce a margin between the clusters of images of a particular identity/class in the embedding space. The embeddings of the same subject/classes are brought close to each other and that of other subjects are pushed away from each other in the output embedding space of the deep CNN model. In addition to that, the loss function ensures that the model produces heterogeneity aware embeddings. Experiments are performed on three popular periocular databases and comparison with existing algorithms  demonstrate state-of-the-art results. The remaining paper is arranged as follows: Section \ref{sec:proposed} contains details of the proposed algorithm. The database used and experiment protocols are discussed in Section \ref{sec:dbase} while the results are discussed in Section \ref{sec:results}.


\section{Proposed Algorithm}
\label{sec:proposed}
\begin{figure*}
  \centering
  \includegraphics[width=0.75\textwidth]{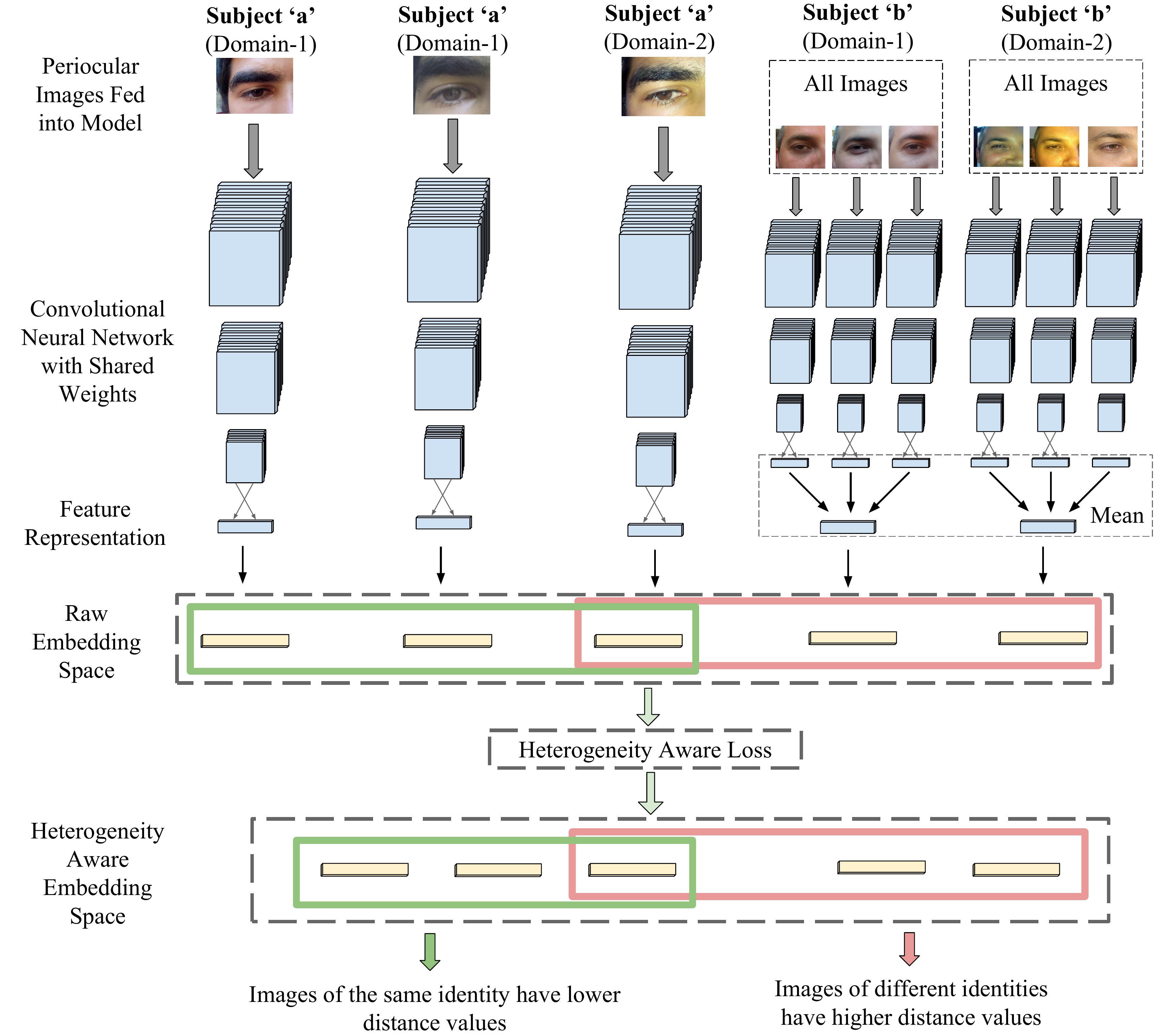}

  \caption{Training the proposed model: periocular images pertaining to different identities are utilized to forward pass through the deep CNN model with shared weights. During training, the loss function (Figure~\ref{fig:hetero_loss}) optimizes the feature representations so that the ones of the same identity are close to each other (i.e. reduce intra-class variations) while others are pushed further apart in the output embedding space of the deep CNN model. $a$ and $b$ refer to different subjects and domain 1 and 2 refer to different image capture scenarios such as indoor/outdoor and with flash/without flash.}
    \vspace{-12pt}
  \label{fig:algorithm}
\end{figure*}

In mobile periocular recognition, heterogeneity may occur due to illumination variations, change in subject to camera distances, and sensor variations. In this section, we illustrate a novel periocular recognition algorithm which trains a deep convolutional neural network model using the proposed heterogeneity aware loss metric. This results in a highly discriminative model producing heterogeneity aware embeddings suitable for matching periocular images captured in unconstrained scenarios. Figure~\ref{fig:algorithm} illustrates the steps involved in the proposed pipeline.  


\subsection{Motivation}
In the homogeneous/ideal scenarios, the vanilla Triplet Loss \cite{schroff2015facenet} can be used which enforces a margin $\alpha$ on the embeddings for a given set of three images known as a triplet. Let $(I_{a1}, I_{a2}, I_{b1})$ be a triplet where $I_{a1}$ is the anchor image of identity/class `$a$', $I_{a2}$ is the positive image which belongs to the same person (identity/class `$a$') and $I_{b1}$ is a negative sample of identity/class `$b$'. Let $g(I_x)$ be the feature embeddings of image $I_x$ and $\tau$ is the set of all triplets and $[z]_+$ is $max(0,z)$. The Triplet loss \cite{schroff2015facenet} aims to minimize the following:

\begin{equation}
  \label{eqn:basic_triplet}
	\sum_{\forall T \in \tau} \left [\left \| g(I_{a1}) - g(I_{a2}) \right \|^2_2 - \left \| g(I_{a1}) - g(I_{b1}) \right \|^2_2   + \alpha \right]_+
\end{equation}
\begin{equation*}
  \forall (I_{a1}, I_{a2}, I_{b1}) \in \tau
\end{equation*}

For a model to produce heterogeneity aware embeddings, it should learn to discriminate between images of different identities as well as bring closer the embeddings of similar identities even in the presence of domain variation at the image level. Such a model should not work with just a single negative sample in the particular triplet. Instead, if the model learns to differentiate between an image of `$a$' and every image of `$b$' (here `$a$' and `$b$' are two different identities) then the model generalizes better because it has to enforce a margin with all the embeddings of the negative class as opposed to a single negative sample.


In order to represent all the embeddings of the negative class, mean embedding of the negative class can be incorporated in the vanilla triplet loss. This means that essentially the centroid of the cluster of images of a negative class is separated from the positive class images. The loss function for the same is as follows:
\vspace{-5pt}
\begin{multline}
  \label{eqn:new_triplet}
  L = \\
  \left [\left \| g(I_{a1}) - g(I_{a2}) \right \|^2_2 - \left \| g(I_{a1}) - \frac{\sum_{i=1}^kg(I_{bi})}{k} \right \|^2_2   + \alpha_1 \right]_+
\end{multline}
where, $I_{a1}, I_{a2}$ belong to class `$a$' and $I_{bi}$ is the $i^{th}$ image of class `$b$' ($I_{a1}$ serves as the anchor),  $\frac{\sum_{i=1}^kg(I_{bi})}{k}$ represents the mean of all the embeddings of a random negative identity `$b$'.

\subsection{Heterogeneity aware embedding space}
Equation \ref{eqn:new_triplet} only incorporates mean embeddings in the same domain and there is no factor of domain/covariate variations. In order to incorporate domain/covariate variation in the model, images needs to be added from different domains for both identities `$a$' and `$b$'. 

Let $p$ and $q$ be the factors of domain variation which we want to incorporate together in the model. Equation \ref{eqn:new_triplet} with the covariate can be expressed as:
\vspace{-15pt}
\begin{multline}
  \label{eqn:cov_new_triplet}
	L_{1} = \\
    \left [\left \| g(I_{a1}^p) - g(I_{a2}^p) \right \|^2_2 - \left \| g(I_{a1}^p) - \frac{\sum_{i=1}^kg(I_{bi}^p)}{k} \right \|^2_2   + \alpha_1 \right]_+
\end{multline}
For multiple domains, we would still like to minimize the distance between the embeddings for same identities and increase it for different identities. This implies, minimize $|| g(I_{a1}^p) - g(I_{a3}^q) ||_2^2$ and maximize $|| g(I_{a1}^p) - g(I_{b1}^q) ||_2^2 $ where $I_x^q$ is an image in different domain and $g(I_x^q)$ is its respective deep CNN model embedding. This means that the cluster of embeddings of a particular class is essentially shrunk as the embeddings are brought closer while the centroid of the cluster of a negative class is pushed away in the embedding space. Hence, the loss equation to train a domain invariant representation can be expressed as:
\begin{multline}
  \label{eqn:hetro_triplet}
	L_{2} = \\
    \left [\left \| g(I_{a1}^p) - g(I_{a3}^q) \right \|^2_2 - \left \| g(I_{a1}^p) - g(I_{bi}^q) \right \|^2_2   + \alpha_2 \right]_+
\end{multline}
\begin{figure}[t!]
  \centering
  \includegraphics[width=1\columnwidth]{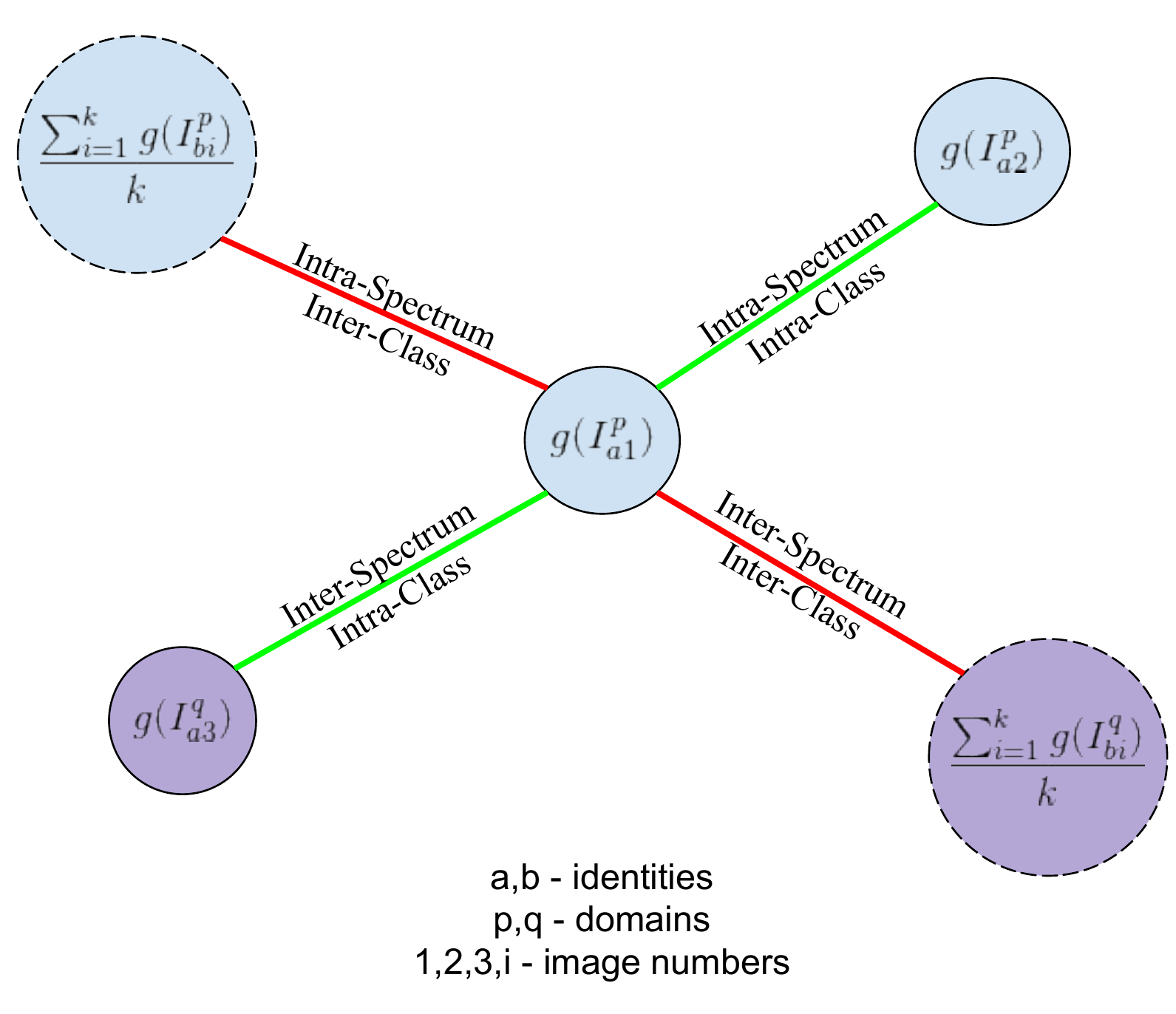}
    \vspace{-8pt}
  \caption{Illustrating the proposed heterogeneity aware loss}
    \vspace{-8pt}
  \label{fig:hetero_loss}
\end{figure}
Representing the negative class by the mean embedding, Equation \ref{eqn:hetro_triplet} can be expressed as:
\begin{equation}
  \label{eqn:cov_hetro_triplet}
	L_{2} =  \left [\left \| g(I_{a1}^p) - g(I_{a3}^q) \right \|^2_2 - \left \| g(I_{a1}^p) - \frac{\sum_{i=0}^kg(I_{bi}^q)}{k}  \right \|^2_2   + \alpha_2 \right]_+
\end{equation}
The final loss equation for creating heterogeneity aware embedding space would be ($\mathcal{L} = L_{1} + L_{2} $):
\vspace{-20pt}
\begin{multline}
	\label{eqn:final_x_loss}
    \mathcal{L} = \\
    \left[ \left \| g(I_{a1}^p) - g(I_{a2}^p) \right \|^2_2 - \left \| g(I_{a1}^p) - \frac{\sum_{i=1}^kg(I_{bi}^p)}{k} \right \|^2_2   + \alpha_1 \right]_+ \\ 
    + \\
    \left[ \left \| g(I_{a1}^p) - g(I_{a3}^q) \right \|^2_2 - \left \| g(I_{a1}^p) - \frac{\sum_{i=0}^kg(I_{bi}^q)}{k}  \right \|^2_2   + \alpha_2 \right]_+
\end{multline}
This loss function can be used to train a domain invariant representation in a deep CNN model, which can be utilized to train for both homogeneous (same domain) and heterogeneous (cross-domain) scenarios. 

\subsection{Implementation Details}
The CNN architecture used for training is LightCNN29~\cite{wu2018light}. The network consists of 29 convolutional layers with $3 \times 3$ filters. There are 4 pooling layers and the feature representation (embedding) layer is 256 dimensional. 
The optimization of the gradient of the loss function is performed via Adam optimizer \cite{kingma2014adam} at a learning rate of $1e^{-5}$ which is slowly decayed. The values of both the summations in the loss are clipped to have a lower bound of 0. The data to be provided to the CNN is sampled randomly from the data available for training and composed into the required tuple. For the experiments, both $\alpha_1$ and $\alpha_2$ have been set to 0.4. 

\section{Databases and Experimental Protocols}
\label{sec:dbase}
\begin{figure}
  \centering
  \includegraphics[width=1\columnwidth]{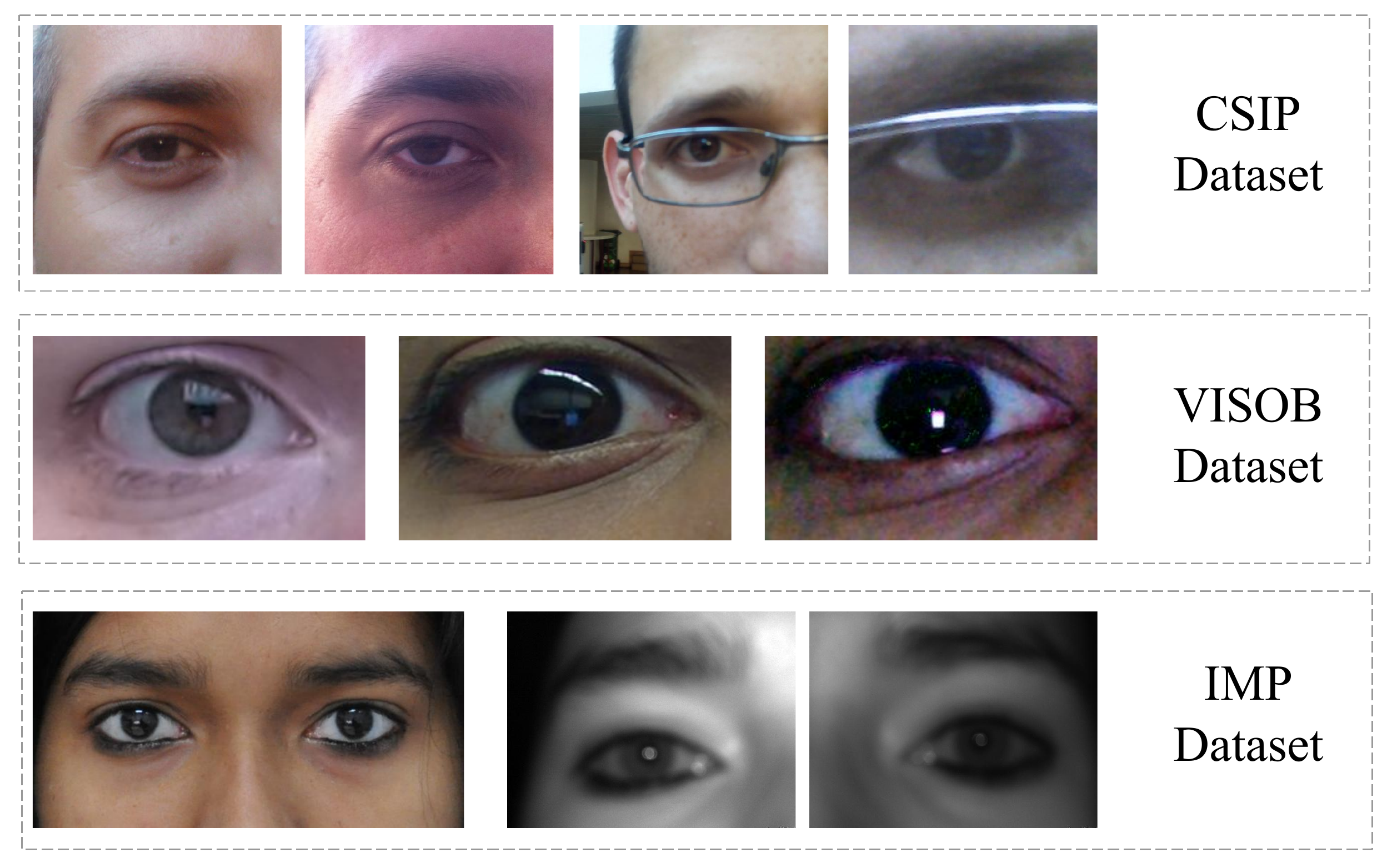}
    \vspace{-8pt}
  \caption{Sample images from the CSIP \cite{santos2015fusing}, VISOB \cite{VISOB_Dataset}, and IMP \cite{sharma2014cross} datasets.}
    \vspace{-12pt}
  \label{fig:eg_imgs}
 \end{figure}

  \vspace{-4pt}
The efficacy of our model is evaluated on two datasets for unconstrained heterogeneous data captured from mobile devices: the CSIP database \cite{santos2015fusing} and the VISOB database \cite{VISOB_Dataset}. Additionally, we have reported results on the IIITD Multi-spectral Periocular Database \cite{sharma2014cross} which has data in different spectrums collected using different sensors including a handheld nightvision camera to show the effectiveness of the proposed algorithm in a general heterogeneous data acquisition scenario. Figure \ref{fig:eg_imgs} shows sample images from these databases.

\subsection{CSIP Database}
  \vspace{-4pt}
The Cross-sensor iris and periocular dataset \cite{santos2015fusing} contains images captured from 4 different mobile phones- Sony Ericsson Xperia Arc, Apple iPhone 4, ThL W200 and Huawei U8510. Images taken from each sensor (mobile phone camera) is further divided into categories denoting front/rear camera and flash/no flash.  The dataset has 2004 right periocular images pertaining to 50 different subjects.
For this dataset, we carry out two experiments, cross sensor and cross illumination periocular recognition. For cross sensor tasks, we train the algorithm on one-vs-all setup, where all images from Apple iPhone 4 serve as one domain, and all images from the remaining sensors are considered as second domain. Training and testing partition is done such that images of subjects 1-40 are used for training and images of subjects from 41-50 form the testing set. 
Additionally we test the proposed algorithm on cross-illumination tasks, such that all the images in the presence of flash form one domain and images captured without flash correspond to different domain. Train test split is similar according to the above protocol.
Results for both the experiments are reported in Tables 1 and 2.



\subsection{VISOB Dataset}
\label{visob}
  \vspace{-4pt}
The VISOB database \cite{VISOB_Dataset} is a large scale dataset from the VISOB ICIP2016 Challenge. It consists of images from 550 subjects captured via the front facing camera of 3 different devices - iPhone 5s, Samsung Note 4 and Oppo N1 in 3 different illumination conditions namely, regular office light, dim light and natural daylight settings. The data was collected in two visits. Only Visit 1 is publicly available. It contains a total of 48,250 images as a part of the enrollment set and 46,797 images as a part of the verification set across all devices and conditions. We perform two experiments on the dataset. \textbf{(a):} In the first experiment, for training, all the images in the enrollment set are used and for testing, the images in the verification set act as probes for the enrolled images via which identification is performed similar to \cite{ahuja2017convolutional}. \textbf{(b):} In order to compare with \cite{zhao2018improving}, the training and testing was performed only on the images captured via the iPhone in day light conditions (as per the protocol used in \cite{zhao2018improving}). 

\subsection{IIITD Multi-spectral Periocular Database}
  \vspace{-4pt}
The IIITD IMP dataset \cite{sharma2014cross} has images captured in three spectrums - visible, near-infrared and night vision, making a total of 1220 images. With 62 subjects in each spectrum and 5 different images corresponding to each subject, the dataset contains 310 images each in the visible and the NIR spectrum. Resolution of the visible spectrum images is $601 \times 301$ and the NIR images are of $540 \times 260$ each. To demonstrate the effectiveness of the proposed approach, no training is performed on this database. The proposed algorithm is evaluated by using the model trained on cropped images of the CASIA NIR-VIS 2.0 face database~\cite{li2013casia}. This is done in order to keep the protocols consistent (to perform comparison) with other cross-spectral periocular recognition methods namely Behera et al~\cite{behera2017periocular} and Ramaiah et al~\cite{ramaiah2016matching}. 
 \vspace{-8pt}
\section{Experimental Results}
\label{sec:results}
  \vspace{-8pt}
The proposed model is evaluated on the datasets discussed in Section \ref{sec:dbase}, and compared with other state-of-the-art algorithms. For CSIP\footnote{ Kandaswamy~\etal~\cite{kandaswamy2017multi} has reported results on this database, but the protocol used in their work is transfer learning based. Santos et al. \cite{santos2015fusing} had performed cross-sensor experiments, but evaluated their algorithm on the entire dataset. Since the proposed method requires training, a direct comparison with \cite{santos2015fusing} is not feasible. Monteiro \etal \cite{monteiro2015comparative} have also computed the results on this dataset, however cross sensor experiments were not performed} dataset, the performance of the proposed algorithm is compared with Triplet Loss \cite{schroff2015facenet} trained in the same way described in Section \ref{sec:dbase}. The training protocol is exactly consistent with the one used for the proposed algorithm. For the cross-illumination and cross-sensor experiments (Table 1 and Table 2) the proposed algorithm achieves a Rank 1 Accuracy of 87.33\% and 89.53\%, respectively. It outperforms \cite{schroff2015facenet} by over 10\% and 5\%, respectively. This illustrates the superiority of the method in generating embeddings that are invariant to the large heterogeneity in the data. Furthermore, apart from the deep learning methods, we also show the comparison with handcrafted features such as Histogram of Oriented Gradients (HOG) \cite{dalal2005histograms} and Daisy features (similar to SIFT) \cite{tola2010daisy}. The results presented in Tables 
1 and 2 corroborate the effectiveness of the proposed model.
\begin{table}
\label{tab:csip_cross_sen}
\caption{Results on the CSIP dataset for cross-sensor mobile periocular recognition tasks.}
  \vspace{-8pt}
\begin{tabular}{|l|c|c|c|}
\hline
\multirow{2}{*}{\textbf{Algorithm}} & \textbf{Identification} & \multicolumn{2}{c|}{\textbf{\begin{tabular}[c]{@{}c@{}}Verification\\ GAR@f FAR\end{tabular}}} \\ \cline{2-4} 
& \textbf{Rank-1(\%)} & \textbf{f=0.1\%} & \textbf{f=10\%} \\ \hline
HOG~\cite{dalal2005histograms} & 62.79 & 2.85& 27.84\\ \hline
DAISY~\cite{tola2010daisy} & 62.40 & 2.49& 33.57\\ \hline
Schroff~\emph{et al.}~\cite{schroff2015facenet} & 84.10 & 12.87& 65.64\\ \hline
Proposed & \textbf{89.53} & \textbf{18.23} & \textbf{75.15} \\ \hline
\end{tabular}

\end{table}

\begin{table}
\label{tab:csip_cross_ilum}
\caption{Results on the CSIP dataset for cross-illumination mobile periocular recognition tasks.}
  \vspace{-8pt}
\begin{tabular}{|l|c|c|c|}
\hline
\multirow{2}{*}{\textbf{Algorithm}} & \textbf{Identification} & \multicolumn{2}{c|}{\textbf{\begin{tabular}[c]{@{}c@{}}Verification\\ GAR@f FAR\end{tabular}}} \\ \cline{2-4} 
& \textbf{Rank-1(\%)} & \textbf{f=0.1\%} & \textbf{f=10\%} \\ \hline
HOG~\cite{dalal2005histograms} & 73.85&3.19 & 27.21\\ \hline
DAISY~\cite{tola2010daisy} & 57.26 & 3.42& 29.80\\ \hline
Schroff~\emph{et al.}~\cite{schroff2015facenet} & 77.42 & 10.17& 59.66 \\ \hline
Proposed & \textbf{87.33} & \textbf{14.53} & \textbf{83.19} \\ \hline
\end{tabular}
  \vspace{-8pt}
\end{table}

\begin{figure*}[t!]
\centering
  \subfloat[IMP Database]{\includegraphics[width=.625\columnwidth]{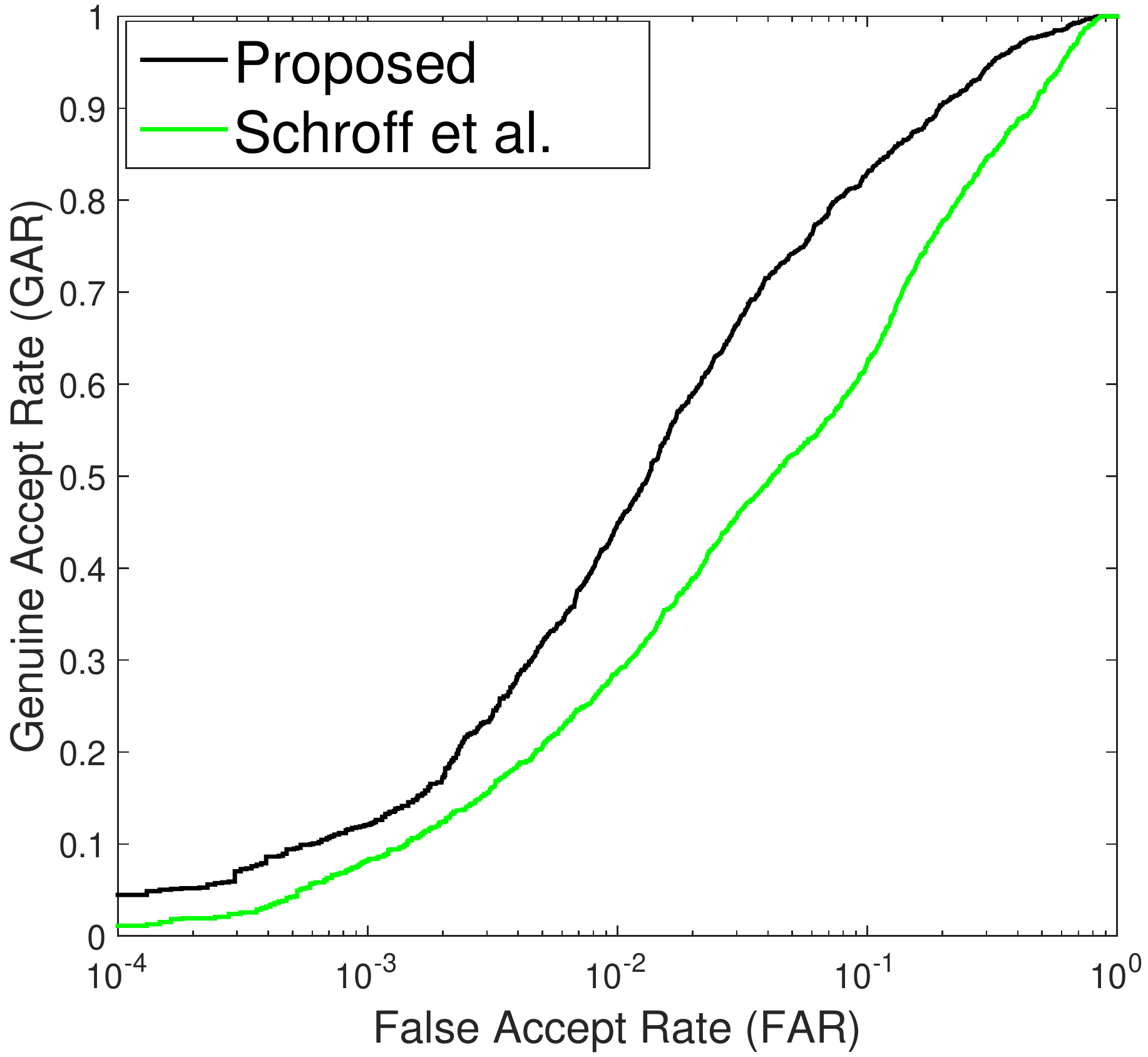}\label{1m}}
  \hspace{2pt}
  \subfloat[CSIP Database (cross illumination)]{\includegraphics[width=.67\columnwidth]{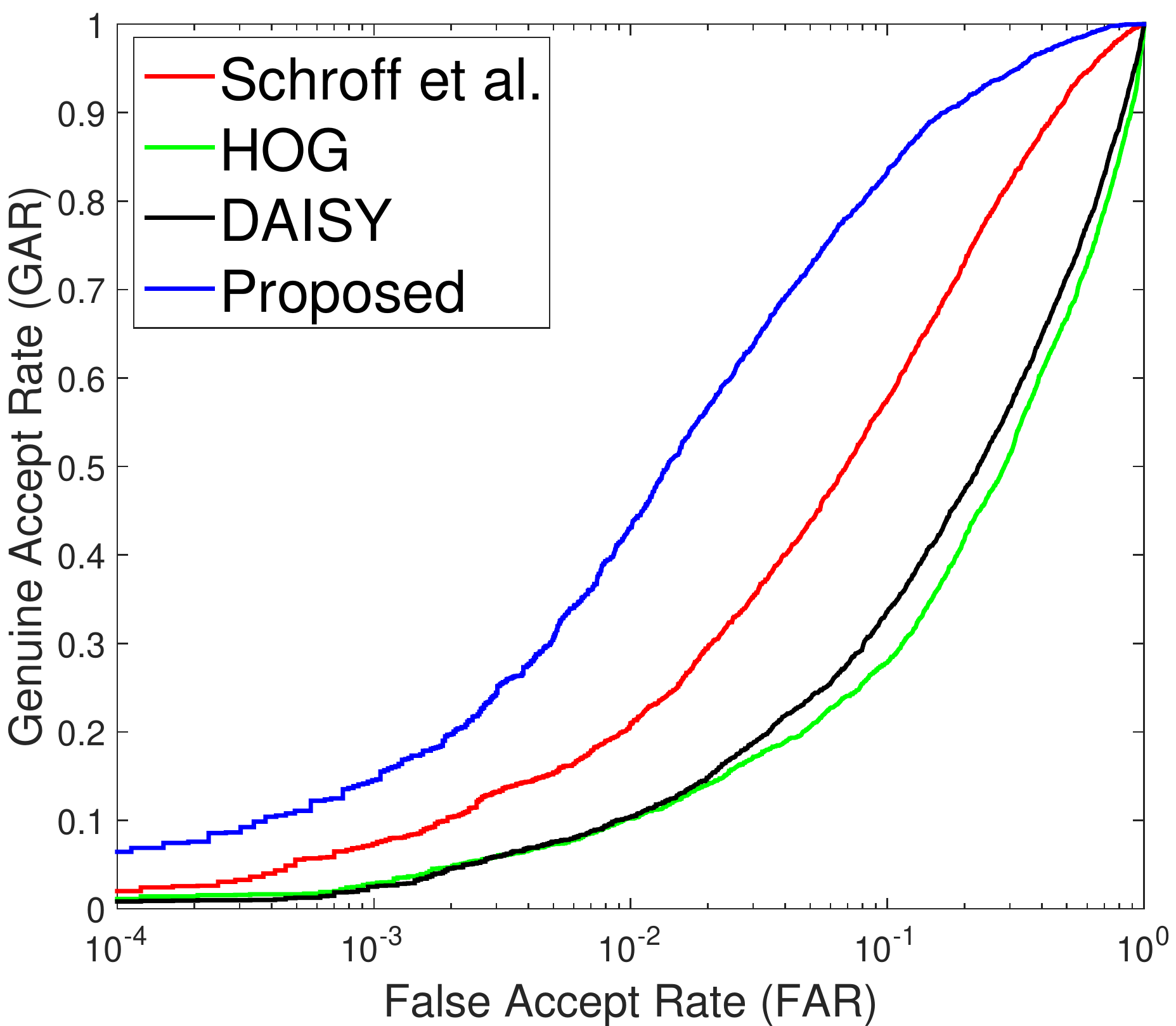}\label{4m}}
   \hspace{2pt}
  \subfloat[CSIP Database (cross sensor)]{\includegraphics[width=.67\columnwidth]{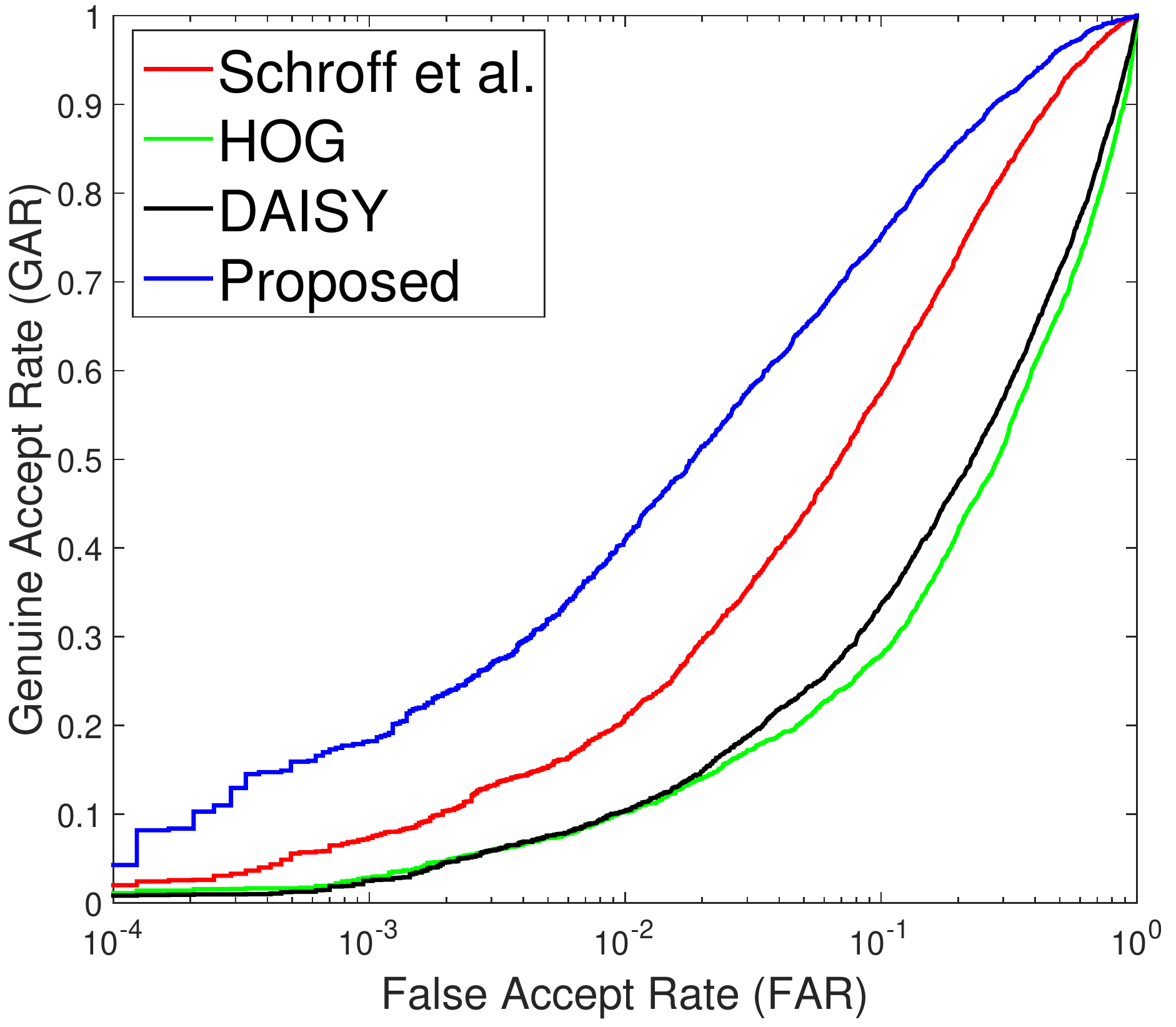}\label{4m}}
 
\caption{ROC curves showing verification accuracies on IMP and CSIP databases}\label{fig:cmc}
  \vspace{-8pt}
   \label{cmc}
  
   \end{figure*}

Table \ref{tab:visob_rank1} summarizes the Rank 1 accuracies of the proposed method on the VISOB Database \cite{VISOB_Dataset} for the experiment \textbf{(a)} (described in section~\ref{visob}). The proposed method outperforms the current state-of-the-art for all devices and lighting conditions, significantly.
Table \ref{tab:visob} 
summarizes the results obtained on the VISOB database for experiment \textbf{(b)}. For comparison with Zhao \etal \cite{zhao2018improving} the same experimental protocol is followed and the results obtained are reported on the same fold. The proposed method obtained an improvement of over $10\%$ over the state-of-the-art EER. 

The results of the IMP dataset are summarized in Table \ref{tab:impres}. It is important to note that no training is performed on this dataset and the reported results are used to illustrate the effectiveness of the model to generate embeddings which can match identities irrespective of the heterogeneity. The method achieves a Genuine Accept Rate of $82.97\%$ at $10\%$ False Accept Rate. As shown in Table 5, the proposed approach  outperforms the state-of-the-art by a very large margin. Results are also compared with the deep learning technique \cite{schroff2015facenet} and the proposed method achieves rank 1 accuracy of $61.2\%$ as compared to $49.36\%$ obtained by \cite{schroff2015facenet}.


\begin{table}

\caption{Rank 1 accuracy on the VISOB Database for experiment 1 with all images.}
\vspace{-4pt}
\scalebox{.75}{
\begin{tabular}{|c|l|c|c|c|c|}
\hline
\multicolumn{2}{|c|}{\multirow{2}{*}{}} & \multicolumn{4}{c|}{\textbf{Rank 1 Accuracy (\%)}}              \\ \cline{3-6} 
\multicolumn{2}{|c|}{}                  & Ahuja \cite{ahuja2017convolutional} & Proposed & Ahuja \cite{ahuja2017convolutional} & Proposed \\ \hline
Phone                    & Condition    & \multicolumn{2}{c|}{Left} & \multicolumn{2}{c|}{Right} \\ \hline
\multirow{3}{*}{Samsung} & Office & 90.45 & \textbf{94.30}    & 91.53  & \textbf{94.71}    \\ \cline{2-6} 
                         & Day    & 92.44 & \textbf{97.15}    & 92.97  & \textbf{98.47}    \\ \cline{2-6} 
                         & Dim    & 93.12 & \textbf{97.19}    & 93.61  & \textbf{98.04}    \\ \hline
\multirow{3}{*}{iPhone}  & Office & 93.54 & \textbf{94.97}    & 93.89  & \textbf{95.88}    \\ \cline{2-6} 
                         & Day    & 95.98 &  \textbf{96.36}    & 94.82  & \textbf{96.06}    \\ \cline{2-6} 
                         & Dim    & 96.09 & \textbf{96.69}    & 96.14  & \textbf{96.54}    \\ \hline
\multirow{3}{*}{Oppo}    & Office & 90.79 & \textbf{91.55}    & 90.23  & \textbf{90.75}    \\ \cline{2-6} 
                         & Day     & 94.21 & \textbf{97.66}    & 94.81  & \textbf{97.25}    \\ \cline{2-6} 
                         & Dim     & 96.31 & \textbf{97.28}    & 96.15  & \textbf{97.07}    \\ \hline
\end{tabular}}
\label{tab:visob_rank1}
\end{table}

\begin{table}

\caption{Results on the VISOB Database with iPhone in daylight}
\vspace{-4pt}
\scalebox{1.01}{
\begin{tabular}{|l|c|c|}
\hline
\textbf{Algorithm} & \textbf{Rank 1 Accuracy(\%)} & \textbf{EER (\%)} \\ \hline
Texton \cite{tan2013towards}          & - & 4.80 \\ \hline
PPDM  \cite{smereka2015probabilistic} & - & 5.03 \\ \hline
SCNN \cite{zhao2017accurate}          & - & 3.30 \\ \hline
Zhao \etal \cite{zhao2018improving}   & - & 1.47 \\ \hline
Proposed                     & \textbf{99.41} & \textbf{1.32} \\ \hline
\end{tabular}}
\end{table}

\label{tab:visob}
\begin{table}

\caption{Results on the IMP dataset for cross-spectrum periocular recognition tasks.}
\vspace{-4pt}
\scalebox{.95}{
\begin{tabular}{|l|c|c|c|}
\hline
\multirow{2}{*}{\textbf{Algorithm}} & \textbf{Identification} & \multicolumn{2}{c|}{\textbf{\begin{tabular}[c]{@{}c@{}}Verification\\ GAR@f FAR\end{tabular}}} \\ \cline{2-4} 
& \textbf{Rank-1(\%)} & \textbf{f=0.1\%} & \textbf{f=10\%} \\ \hline
Ramaiah~\emph{et al.}~\cite{ramaiah2016matching}& - & - & 18.35\\ \hline
Behara~\emph{et al.}~\cite{behera2017periocular}& - & - & 25.03\\ \hline
Schroff~\emph{et al.}~\cite{schroff2015facenet} & 49.36 & 8.23 & 62.27\\ \hline
Proposed & \textbf{61.20} & \textbf{12.07} & \textbf{82.97} \\ \hline
\end{tabular}}
\label{tab:impres}
\end{table}

Apart from the accuracies observed, we have made following observations:\\
\textbf{Cross-Database Performance}:
In order to compare the performance of the proposed approach with state-of-the-art algorithms~\cite{behera2017periocular,ramaiah2016matching} for the IMP dataset, we performed testing on this dataset without training on any image of this dataset. The deep CNN model was trained on the CASIA NIR-VIS 2.0 dataset~\cite{li2013casia}. Periocular images were extracted from the face images of this dataset for training. This training was performed with spectrum as the heterogeneity and then the trained model was utilized for testing on the entire IMP dataset. This mimics a cross-database train-test scenario. As shown in Table \ref{tab:impres}, the proposed algorithm produces state-of-the-art results, which shows that our model is generalizable to datasets on which no fine-tuning or training is performed. It should also be noted that the CASIA and IMP datasets contain subjects pertaining to different ethnicities and the images are collected using different sensors. High verification performance with cross-database testing is a strong indication of the generalizability of the algorithm.  \\
\textbf{Hard Mining}:
Most deep metric learning algorithms~\cite{hermans2017defense,schroff2015facenet} are heavily dependent on hard mining of samples for training. However, the proposed method, produces better results than one of the most popular deep metric learning algorithms~\cite{schroff2015facenet} without any hard-mining. This saves a huge amount of training time and is a testament to the efficacy of the proposed algorithm.  \\
\textbf{Testing Time:}
On Intel Core $i7$ workstation with 32GB of RAM and NVIDIA GTX 1080ti GPU, the average time for matching a pair of images is 50.5 microseconds.  
\vspace{-4pt}
\section{Conclusion and Future Research}
\vspace{-4pt}
Mobile periocular recognition requires addressing heterogeneity due to illumination variations, subject-to-camera distances, sensor variations, and indoor-outdoor variations. To address this research challenge, a heterogeneity aware loss is proposed to train deep CNN model which helps in creating domain invariant embedding space. The proposed algorithm for periocular recognition in unconstrained environments achieves state-of-the-art results. Although the results are shown on periocular recognition tasks, the proposed loss metric can also be extended for other recognition tasks such as recognizing faces with disguise variations~\cite{dhamecha2014recognizing,kushwaha2018disguised,singh2009face}, heterogeneous face recognition~\cite{dhamecha2014effectiveness,ghosh2015feature}, and iris/periocular recognition with multiple cameras or covariates~\cite{arora2012iris,keshari2016mobile}. 
\vspace{-8pt}
\section{Acknowledgement}
\vspace{-4pt}
M. Vatsa and R. Singh are partly supported by Infosys Center for Artificial Intelligence, IIIT Delhi. S. Ghosh is partly supported through TCS PhD Fellowship.

{\small
\bibliographystyle{ieee}
\bibliography{submission_example}
}

\end{document}